\documentclass[letterpaper,10pt,conference]{ieeeconf}
\IEEEoverridecommandlockouts
\usepackage{algorithm}

\usepackage[hidelinks]{hyperref}
\usepackage{multirow}
\usepackage{algpseudocode} 
\usepackage{optidef}
\usepackage{amsmath}
\usepackage{amssymb}
\usepackage{mathtools}
\usepackage{cite}
\usepackage[dvipsnames]{xcolor} 
\usepackage{tabularray} 
\usepackage{graphicx}
\graphicspath{{./images/}}
\DeclareGraphicsExtensions{.pdf,.png,.jpg}
\usepackage{amsmath}
\usepackage{amssymb}
\usepackage{booktabs}
\begin{document}

\title{\LARGE \bf{Improved Postural Stability Using a Lightweight Semi-Active Soft Back Support Device Under Standing Perturbations}}

\author{Rohan Khatavkar, Jiefeng Sun, and Hyunglae Lee\textsuperscript{*}
%
\thanks{The authors are with the School for Engineering of Matter, Transport and Energy, Tempe, AZ 85287, USA. \textsuperscript{*}Corresponding author: Hyunglae~Lee (hyunglae.lee@asu.edu)}%
}

\maketitle
\begin{abstract}
Older adults are particularly susceptible to falls following perturbations during standing, such as forward loss of balance. Back support devices that assist trunk extension may help mitigate fall risk by preventing excessive trunk flexion. Previous studies have investigated heavy back support devices; however, these systems often introduced adverse effects on stability due to their added mass, which shifted the body’s natural center of mass unfavorably. In contrast, lightweight passive devices have shown limited benefits, as they can generate only modest assistive forces during the relatively small trunk flexion associated with forward balance loss. In this study, we evaluated the effects of a lightweight semi-active soft back support device on postural stability following standing perturbations. Our device combines an active element (a pneumatic artificial muscle) in parallel with a passive elastic band. The active element rapidly provides assistive force following a perturbation, overcoming the limitations of passive devices. Experiments conducted with five healthy individuals demonstrated that the semi-active device significantly reduced whole-body angular momentum and increased the margin of stability, indicating improved balance recovery performance. These results highlight the promise of semi-active soft wearable robots as an effective and lightweight strategy for fall prevention during standing perturbations.
\end{abstract}

\section{Introduction}
The elderly population is at a high risk of falls under perturbations \cite{norgaard_effect_2023,ambrose_risk_2013}. The elevated fall risk is associated with delayed reactive stepping and age-related muscular weakness \cite{tashiro_effect_2024, okubo_stepping_2021}. One of the most common perturbations is a forward loss of balance (e.g., consider an older adult standing in a moving train; the person would experience a forward loss of balance if the train suddenly stops) \cite{dominguez_postural_2020}. To recover from such a trip, humans must step forward and activate the back extensor muscles to prevent excessive trunk flexion \cite{graham_muscle_2014, van_der_burg_out--plane_2005}. It is significantly more challenging for older adults to prevent trunk flexion or to rapidly flex their hip (to step forward), compared to young adults \cite{dominguez_postural_2020}. As a result, devices that prevent trunk flexion (or assist trunk extension) or assist hip flexion have the potential to improve balance among the elderly population.

The simplest devices that prevent excessive trunk flexion include canes and walkers. However, these devices require the user to actively move them while reacting to a perturbation. Since older adults have diminished reflex actions, such devices may not be effective \cite{gell_mobility_2015}. On the other hand, exoskeletons providing assistive force for trunk or hip motion have the potential to improve stability following perturbations. Hip exoskeletons can improve reactive stability following perturbations, depending on controller timing and device mass \cite{tagliaferri_systematic_2026,leestma_dynamic_2024}. However, because hip exoskeletons primarily generate torque about the hip joint, their influence on upper-body dynamics and trunk stabilization is indirect. In some cases, poorly tuned assistance or added distal mass can even interfere with natural recovery strategies, such as rapid stepping or trunk stabilization \cite{normand_effect_2023}.

\begin{figure}[t!]
    \centering
    \includegraphics[width = 1\linewidth]{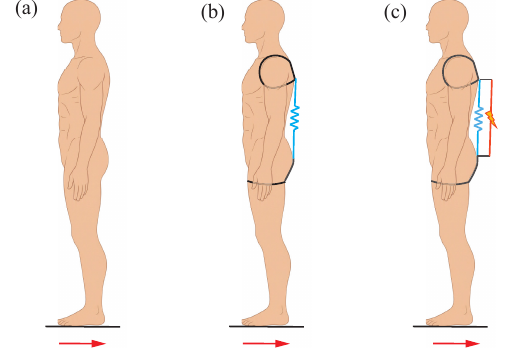}
    \caption{(a) Standing perturbation with no back support device. (b) Standing perturbation with a passive back support device. (c) Standing perturbation with a semi-active back support device.}
    \label{fig: First}
    \vspace{-0.75em}
\end{figure}

\begin{figure*}[ht!]
    \centering
    \includegraphics[width = 1\linewidth]{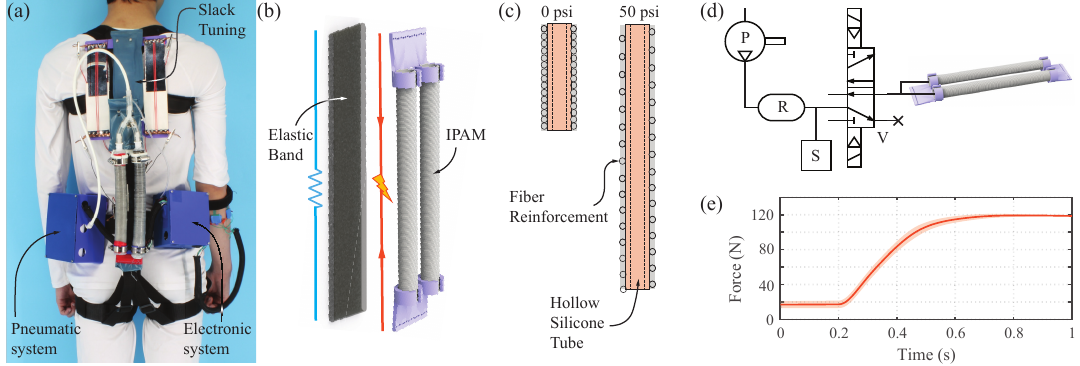}
    \caption{(a) A lightweight semi-active soft back support device. (b) Parallel arrangement of a passive element (elastic band) with an active element (Inverse Pneumatic Artificial Muscle, IPAM). (c) Working principle of the IPAM: It extends axially when pressurized; it contracts when deflated, thereby generating an active force. (d) An untethered pneumatic system to inflate and deflate the IPAM. (e) Rapid increase in active force enabled by rapid IPAM deflation through the pneumatic system. Force was measured using a force testing machine. The shaded area shows $\pm$1 standard deviation over 3 trials.}
    \label{fig: device design}
    \vspace{-0.75em}
\end{figure*}

On the other hand, back support devices (BSDs) act directly on trunk extension moments, suggesting that they may be particularly effective at resisting perturbations that produce excessive trunk flexion. Biomechanics researchers have evaluated the effect of commercially available passive BSDs like Laevo Flex (\url{en.laevo.nl}), BackX (\url{www.suitx.com}), and HeroWear Apex (\url{https://herowearexo.com}) on stability. However, several studies report insignificant or adverse effects of using these devices on stability during quiet stance, forward loss of balance, and trip- and slip-like perturbations \cite{park_effects_2021, park_wearing_2024}. These effects, which are contrary to expectations, may be partly attributed to the weight of these devices (2.8-4.3 kg) \cite{park_effects_2021}. Wearing a heavy device on the body can significantly shift the CoM and impair stability, as users may be unfamiliar with the shifted CoM. A lightweight (1.4 kg) passive BSD, HeroWear Apex, improved the step speed following a trip-like perturbation. However, it only marginally improved other balance-related metrics, such as trunk angular velocity \cite{dooley_occupational_2024}. The major reason for observing only marginal improvements with this lightweight device may be the relatively low force applied by passive devices when the perturbation causes a small trunk flexion angle. Though a passive device with high stiffness can resolve this problem, such high stiffness may restrict the flexion range of motion and lead to discomfort during other voluntary back movements.

Using an active BSD can ensure sufficient force at small trunk flexion angles during perturbations and maintain transparency during voluntary flexion. However, most state-of-the-art active BSDs are heavy (2.7-9.2 kg) and bulky, and may impair stability \cite{kermavnar_effects_2021, golriz_effect_2015}. A previously developed light-weight semi-active BSD could potentially improve stability by virtue of its low weight (1.97 kg) and its ability to provide adequate force at small trunk flexion angles \cite{khatavkar_soft_2026}.

In this work, we test the effect of using a lightweight semi-active BSD on stability following a standing perturbation. The device combines a passive elastic element in parallel with an active element, enabling it to provide adequate assistive force while maintaining a lightweight, compact design. We evaluated the effect of the device using only the passive element (Passive Mode) and both the passive and active elements together (Active Mode) on stability following standing perturbations (Fig. \ref{fig: First}). We quantify stability using whole body angular momentum (WBAM) and margin of stability (MOS), which are two common metrics used for stability quantification \cite{leestma_linking_2023, hof_condition_2005}. We hypothesize that the root mean squared (RMS) WBAM and WBAM range would decrease with the device compared to the no-device condition. We also hypothesize that the minimum MOS would increase with the device compared to the no-device condition.




\section{Methods}
\subsection{Semi-active Back Support Device}
The device (shown in Fig. \ref{fig: device design}(a)) combines a passive elastic band with an active element in parallel, as shown in Fig. \ref{fig: device design}(b). The parallel arrangement allows the device to apply adequate total assistive force in a compact and lightweight design. The elastic band applies a passive force between the upper back and the thigh attachments as it elongates with trunk flexion. The active element (inverse pneumatic artificial muscle, IPAM) augments the passive force on demand.

\begin{figure}[b!]
\vspace{-0.75em}
    \centering
    \includegraphics[width = 1\linewidth]{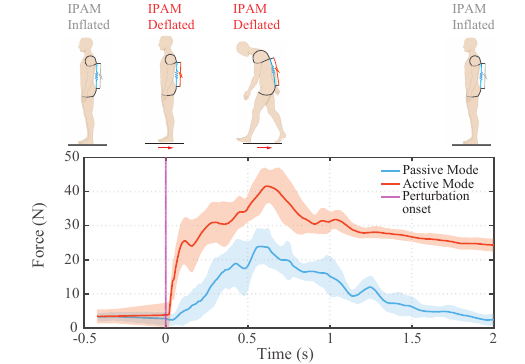}
    \caption{Device force profiles measured by an onboard load cell. Shaded areas show $\pm$1 standard deviation over 15 trials for a participant.}
    \label{fig: exp protocol v2}
\end{figure}

Unlike conventional pneumatic muscles that contract when pressurized, the IPAM elongates under pressure and contracts upon deflation, as shown in Fig. \ref{fig: device design}(c) \cite{hawkes_design_2016}. A lightweight, portable pneumatic system actuates the IPAM (Fig. \ref{fig: device design}(d)). Three air reservoirs (R, 115.53\,cc each; x-small, Robart) pressurized by a piston pump (P1, BD-07A38L-12V, Bodenflo; 455\,g, 75\,dB) serve as the compressed air source for the IPAMs. A 5-port, 2-way solenoid valve (V, Airtac) controls the IPAM actuation. The IPAM is pressurized to 50 psi when the valve V is turned on, and it deflates when the valve is turned off. A pressure sensor (S, 4525-SS5A100GP, TE Connectivity) measures the reservoir pressure to charge it up to 60 psi when required. The portable electronic system uses a primary microcontroller (Feather 32u4, Adafruit) to actuate the pneumatic components accordingly. The piston pump (P) is controlled via an electromechanical relay. The solenoid valve V is switched using a transistor array (TBD62083APG-ND, Toshiba). Power for the pneumatic system is provided by an 11.1\,V, 5200\,mAh battery, which supplies the piston pump and the valve. A 3.3\,V buck converter supplies power to the Feather 32u4.

Rapid deflation is crucial for generating adequate assistive force following a perturbation. Though we use a compact, portable pneumatic actuation system, the IPAM can deflate rapidly, as shown by a force test (using \#3367, Instron) in Fig. \ref{fig: device design}(e). In the perturbation experiment, the IPAM is inflated in the upright standing posture, as shown  in Fig. \ref{fig: exp protocol v2}. At the onset of a perturbation, the IPAM deflates to rapidly add force (measured using an onboard load cell, LCM300, FUTEK). This rapid rise in force provides adequate assistance following the perturbation onset, even though the trunk flexion angle is small. We denote such an assistive profile as the ``Active Mode''.

In this study, we also evaluated a ``Passive Mode'' as a baseline condition. In the Passive Mode, we only use the elastic band, which has a stiffness of 0.8 N/mm (a typical value for passive devices) \cite{lamers_feasibility_2018}. Thus, the passive mode was equivalent to using a passive device. In the Passive Mode, the assistive force vs. time profile is shown in Fig. \ref{fig: exp protocol v2}. The force does not increase rapidly following the perturbation onset, as it is directly proportional to the trunk flexion angle.

\subsection{Participants}
Five healthy young adults (age: $22.6\pm2.3$ years, height: $1.75\pm0.08$ m, weight: $73.0\pm11.7$ kg) were recruited for the stability analyses. All experimental protocols were approved by the Institutional Review Board at Arizona State University (STUDY00020698).

\subsection{Experimental Setup and Protocol}
\begin{figure}[t!]
    \centering
    \includegraphics[width = 0.9\linewidth]{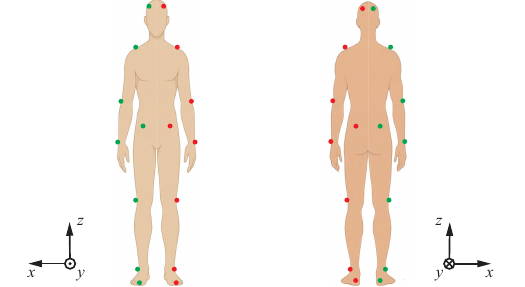}
    \caption{Simplified Markerset}
    \label{fig: markerset}
    \vspace{-0.75em}
\end{figure}

The experiment was conducted on a split-belt force treadmill (Bertec) while an eight-camera optical motion tracking system (Vicon) recorded human motion. A simplified version of the Vicon full body Plug-in AI marker set was used for capturing motion \cite{noauthor_plug-gait_2025}. We included 22 markers, as shown in Fig. \ref{fig: markerset}. Other markers from the full body Plug-in Gait marker that interfered with the device were excluded. A safety harness was used to prevent injuries to the participants in case of a fall during the experiment.

\begin{figure}[t!]
    \centering
    \includegraphics[width = 1\linewidth]{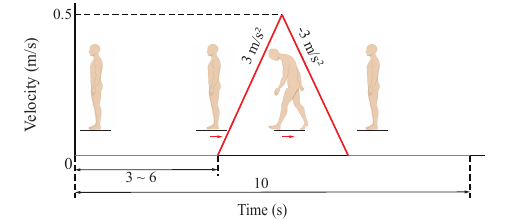}
    \caption{Velocity profile of the treadmill belts.}
    \label{fig: exp protocol}
\end{figure}

Before the experiment, the protocol was explained in detail to the participants, and they signed their written consent. The participants then stood on the treadmill and wore the safety harness. Then, the participants wore the device, and reflective markers were attached to appropriate locations. During the experiment, the subject stood still on the treadmill as the motion capture began. After a random duration between 3 and 6 s, the treadmill moved backward with the velocity profile shown in Fig. \ref{fig: exp protocol}. The subjects were asked to react naturally to the perturbation with the goal of regaining balance. Each trial lasted 10 s. Each participant completed 15 such trials in each of the three conditions: (1) No device, (2) Device in Passive Mode, and (3) Device in Active Mode. The order of the three device conditions was randomized within subjects and across subjects.

\begin{figure}[b!]
\vspace{-0.75em}
    \centering
    \includegraphics[width=1\columnwidth]{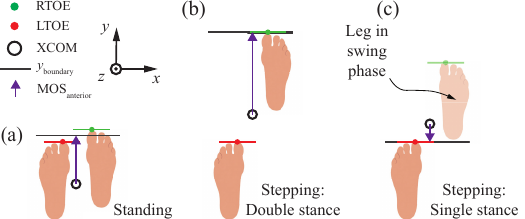}
    \caption{The anterior boundary of support and the anterior margin of stability.}
    \label{fig: BOS}
\end{figure}

\begin{figure*}[thpb]
\centering
\includegraphics[width=1\linewidth]{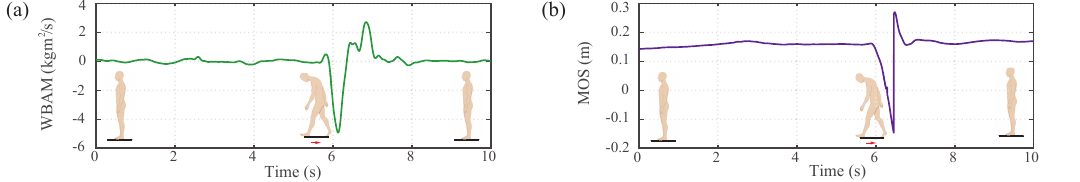}
\caption{(a) WBAM in a representative trial of a representative subject. (b) MOS in a representative trial of a representative subject.}
\label{fig:WBAM_MOS}
\vspace{-0.75em}
\end{figure*}

\subsection{Data Processing}
\subsubsection{Whole-Body Angular Momentum}
Whole-body angular momentum (WBAM) was computed from three-dimensional marker data. The marker data were defined in the following coordinate system: X represented the medio-lateral direction, Y represented the antero-posterior direction, and Z represented the vertical direction. A 14-segment model (pelvis, thorax, head, and left/right humerus, radius, femur, tibia, and foot) was used. Segment masses were derived from total body mass using standard mass fractions in Vicon documentation \cite{noauthor_plug-gait_2025}, and segment lengths and center-of-mass (COM) positions were obtained from anatomical landmarks. The whole-body COM was computed as the mass-weighted sum of segment COMs. Segment linear velocities were obtained by differentiating COM positions using a four-point central-difference scheme; segment angular velocities were computed from the time derivative of the unit vector along each segment's longitudinal axis. Segment moments of inertia were approximated as $I_i = m_i (k_i L_i)^2$, where $m_i$ is segment mass, $L_i$ is segment length, and $k_i$ is the segment radius of gyration.

WBAM was computed as the sum of transfer and local angular momentum \cite{leestma_linking_2023}. The transfer term is
\begin{equation}
\mathbf{L}_{\mathrm{transfer}} = \sum_i m_i (\mathbf{r}_i - \mathbf{r}_{\mathrm{COM}}) \times \mathbf{v}_i,
\end{equation}
where $\mathbf{r}_i$ and $\mathbf{v}_i$ are the position and velocity of the $i$th segment COM, and $\mathbf{r}_{\mathrm{COM}}$ is the whole-body COM. The local term is
\begin{equation}
\mathbf{L}_{\mathrm{local}} = \sum_i I_i \boldsymbol{\omega}_i,
\end{equation}
where $\boldsymbol{\omega}_i$ is the angular velocity of segment $i$. The total WBAM vector, $\mathbf{L}_{\mathrm{body}} = \mathbf{L}_{\mathrm{transfer}} + \mathbf{L}_{\mathrm{local}}$ was converted to SI units (kg$\cdot$m$^2$/s). For analysis, only the sagittal-plane (YZ plane) angular momentum, i.e., the \emph{x}-component $L_x$ was used. For each trial, we computed the range of $L_x$ (maximum minus minimum) and the RMS of $L_x$ over the trial. Marker data were low-pass filtered at 6~Hz (fourth-order Butterworth) before differentiation. Fig. \ref{fig:WBAM_MOS}(a) shows WBAM during a representative trial of a representative subject.

\subsubsection{Margin of Stability}
The margin of stability (MOS) was computed using the extrapolated center of mass (XCOM) in the anterior-posterior direction (\emph{y}-component) ~\cite{hof_condition_2005}. The whole-body COM position and velocity were obtained from the same segmental model as for WBAM. The anterior-posterior XCOM was computed as
\begin{equation}
\mathrm{XCOM}_y = y_{\mathrm{COM}} + \frac{\dot{y}_{\mathrm{COM}}}{\sqrt{g/h}},
\end{equation}
where $y_{\mathrm{COM}}$ and $\dot{y}_{\mathrm{COM}}$ are the COM position and velocity in the anterior direction, $g$ is gravitational acceleration, and $h = 1.2\ell$ with $\ell$ denoting the participant's leg length. Stance was detected from the vertical ground reaction force; a limb was considered in stance when the magnitude of the vertical force exceeded 4.5\% of body weight. The anterior boundary of the base of support was defined at each time instant as the anterior-posterior position of the leading toe (left or right) of the stance foot or, in double stance, the more anterior toe (Fig. \ref{fig: BOS}). The anterior MOS was defined as the signed distance from the XCOM to that boundary:
\begin{equation}
\mathrm{MOS}_{\mathrm{anterior}} = y_{\mathrm{boundary}} - \mathrm{XCOM}_y.
\end{equation}
Positive values indicate that the XCOM has not yet reached the boundary. For each trial, we extracted the minimum MOS over the analyzed interval as the primary stability metric. Force data were low-pass filtered at 6~Hz and downsampled to match the marker sampling rate (200~Hz) for synchronization.

\subsection{Statistical Analyses}
\label{sec:statistical}

For individual results, metrics were reported without normalization: WBAM in kg\,$\cdot$\,m$^2$/s and minimum MOS in cm. For group-level comparisons across conditions (No Device, Passive Mode, Active Mode), normalized metrics were used: WBAM was normalized by the product of body mass and height (kg\,$\times$\,m), and minimum MOS by height (m), using subject-specific values. 

A linear mixed-effects model (LME) was fitted to the full trial-level data, with condition as a fixed effect and subjects as a random effect. One LME was fitted per metric (WBAM range, WBAM RMS, and minimum MOS in the anterior-posterior direction). The overall effect of condition was assessed with an $F$-test on the fixed effect; pairwise comparisons between conditions were performed using contrast tests on the fixed-effect coefficients, with a Bonferroni correction for the three comparisons. Significance was set at $\alpha = 0.05$, and results were summarized with brackets and asterisks on group bar plots ($^*p < 0.05$, $^{**}p < 0.01$, $^{***}p < 0.001$).

\section{Results}
\subsection{Individual Results}
\subsubsection{WBAM RMS}
Table \ref{tab:wbam_rms} shows the RMS values of WBAM for individual subjects across the three experimental conditions. The WBAM RMS decreased slightly with the device in Passive Mode in four of the five subjects, while for one subject (S03) it increased in the Passive Mode. These observations suggest that a passive BSD may have limited effectiveness in reducing WBAM. In the Active Mode, the WBAM RMS was consistently lower compared to both the No Device condition and the Passive Mode across all five subjects. Thus, the addition of the active assistance at the onset of the perturbation appears to play an important role in reducing WBAM RMS.

\subsubsection{WBAM Range}
Table \ref{tab:wbam_range} presents the WBAM range across the three experimental conditions for individual subjects. Consistent with the WBAM RMS results, the WBAM range decreased slightly in four of the five subjects in the Passive Mode compared to the No Device condition, while it increased for subject S03. Thus, the WBAM range analyses also suggest that passive BSDs may have limited effectiveness in reducing WBAM during forward loss of balance. In contrast, the WBAM range in the Active Mode was consistently lower than in both the No Device condition and the Passive Mode across subjects. Therefore, the WBAM range analysis further supports the importance of providing active assistance at the onset of the perturbation to reduce the user’s WBAM.

\begin{table}[t!]
\centering
\caption{WBAM RMS}
\label{tab:wbam_rms}
\vspace{-0.5em}
\begin{tabular}{lccc}
\toprule
Subject & No device & Passive Mode & Active Mode \\
\midrule
S01 & $1.41 \pm 0.29$ & $1.32 \pm 0.08$ & $1.01 \pm 0.27$ \\
S02 & $2.11 \pm 0.21$ & $1.79 \pm 0.14$ & $1.76 \pm 0.23$ \\
S03 & $1.25 \pm 0.13$ & $1.43 \pm 0.20$ & $0.86 \pm 0.18$ \\
S04 & $1.31 \pm 0.15$ & $1.11 \pm 0.06$ & $0.62 \pm 0.14$ \\
S05 & $1.64 \pm 0.17$ & $1.40 \pm 0.01$ & $0.94 \pm 0.19$ \\
\bottomrule
\end{tabular}
\footnotesize{\\Values are reported as mean $\pm$ standard deviation in kg\,$\cdot$\,m$^2$/s.}
\vspace{-0.5em}
\end{table}

\begin{table}[t!]
\centering
\caption{WBAM range}
\label{tab:wbam_range}
\vspace{-0.75em}
\begin{tabular}{lccc}
\toprule
Subject & No device & Passive Mode & Active Mode \\
\midrule
S01 & $11.00 \pm 3.28$ & $11.15 \pm 1.57$ & $8.17 \pm 2.78$ \\
S02 & $15.93 \pm 1.03$ & $14.10 \pm 0.98$ & $12.85 \pm 1.57$ \\
S03 & $10.42 \pm 1.30$ & $11.42 \pm 1.94$ & $7.75 \pm 1.72$ \\
S04 & $9.88 \pm 0.50$ & $8.80 \pm 0.78$ & $4.90 \pm 1.34$ \\
S05 & $12.42 \pm 1.66$ & $11.13 \pm 0.87$ & $7.21 \pm 1.77$ \\
\bottomrule
\end{tabular}
\footnotesize{\\Values are reported as mean $\pm$ standard deviation in kg\,$\cdot$\,m$^2$/s.}
\vspace{-0.5em}
\end{table}

\begin{table}[t!]
\centering
\caption{Minimum MOS}
\label{tab:min_mos}
\vspace{-0.75em}
\begin{tabular}{lccc}
\toprule
Subject & No device & Passive Mode & Active Mode \\
\midrule
S01 & $-0.31 \pm 1.10$  & $0.535 \pm 1.16$   & $1.12 \pm 1.08$ \\
S02 & $-5.64 \pm 0.77$  & $-6.42 \pm 0.60$  & $-6.46 \pm 1.42$ \\
S03 & $-9.97 \pm 1.31$ & $-13.62 \pm 3.17$ & $-8.82 \pm 1.91$ \\
S04 & $-8.77 \pm 0.74$  & $-11.85 \pm 3.39$ & $-0.177\pm 1.40$ \\
S05 & $-7.55 \pm 1.40$ & $-10.08 \pm 0.90$  & $-1.47 \pm 1.59$ \\
\bottomrule
\end{tabular}
\footnotesize{\\Values are reported as mean $\pm$ standard deviation in cm.}
\vspace{-0.5em}
\end{table}

\subsubsection{Minimum MOS}
Table \ref{tab:min_mos} presents the minimum MOS values across the three experimental conditions for individual subjects. Most entries are negative because subjects responded to the perturbations by stepping forward. During the swing phase, the anterior boundary of support is located posterior to the body’s CoM, resulting in negative MOS values. In the Passive Mode, the minimum MOS was generally lower compared to the No Device condition, suggesting a potential detrimental effect of the passive BSD on stability. In contrast, the minimum MOS in the Active Mode was generally higher than in the No Device condition, indicating a beneficial effect of the active assistance applied at perturbation onset. Fig. \ref{fig:WBAM_MOS}(b) shows the MOS trajectory during a representative trial for one subject.

\subsection{Group Average Results}
In this section, we describe the results of comparing the three experimental conditions across subjects.
\subsubsection{WBAM RMS}
\begin{figure*}[thpb]
\centering
\includegraphics[width=0.95\linewidth]{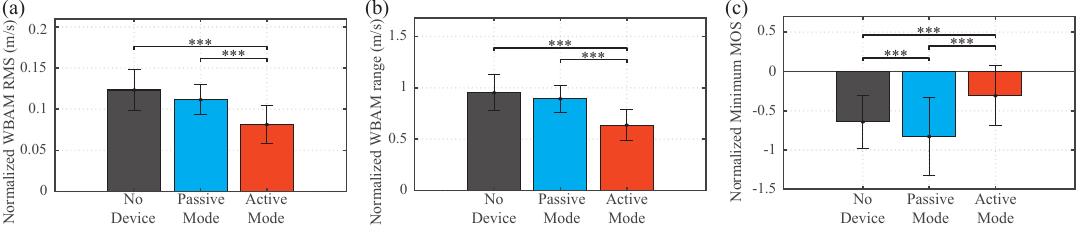}
\caption{(a) Group average WBAM RMS normalized by the product of subject-specific weight and height. (b) Group average WBAM range normalized by the product of subject-specific weight and height. (c) Group average minimum MOS normalized by the subject-specific weight.}
\label{fig:reults}
\end{figure*}

Fig. \ref{fig:reults}(a) shows the RMS WBAM across subjects in the three experimental conditions. It can be observed that the WBAM RMS reduced with the device in Passive Mode only marginally ($p = 0.071$). Consistent with the individual results, these group average results also indicate that a passive BSD may not be efficacious in significantly reducing WBAM. Similar to the individual results, the WBAM RMS was significantly lower ($p < 0.001$) in the Active Mode compared to both the No Device condition and the Passive Mode. Thus, the addition of active force at the onset of the perturbation was essential in reducing WBAM RMS.

\subsubsection{WBAM Range}
Fig. \ref{fig:reults}(b) shows the WBAM range in the three experimental conditions across subjects. The WBAM range did not change significantly in the Passive Mode compared to the No Device condition ($p = 0.109$). Thus, the group average range analyses also imply that passive BSDs may have limited effectiveness in reducing WBAM during forward loss of balance. Similar to the individual results, the WBAM range was significantly ($p < 0.001$) lower in the Active Mode compared to both the No Device condition and the Passive Mode. Therefore, the WBAM range analyses further reinforce the necessity of active force following the onset of the perturbation in effectively reducing WBAM.

\subsubsection{Minimum MOS}
Fig. \ref{fig:reults}(c) shows the analyses of minimum MOS in the three experimental conditions across subjects. The minimum MOS was significantly lower ($p < 0.001$) in the Passive Mode compared to the No Device condition. This indicates a detrimental effect of the passive BSD on minimum MOS. On the other hand, the minimum MOS increased significantly ($p < 0.001$) in the Active Mode compared to both the No Device condition and the Passive Mode. The results suggest a potential beneficial effect of the active force applied at perturbation onset.

\section{Discussion}

In this paper, we evaluated the effect of a lightweight semi-active BSD on stability following a standing perturbation. WBAM and MOS were evaluated in two conditions of the device: (1) Passive Mode, equivalent to a passive BSD, and (2) Active Mode, in which an active element rapidly adds force to the passive force at the onset of a perturbation. We showed that the WBAM RMS and range were reduced significantly only in the Active Mode compared to the No Device condition and the Passive Mode. The Passive Mode was not effective, likely because the device was unable to apply adequate force at small trunk flexion angles following the perturbation. On the other hand, IPAM (active element) deflation in the Active Mode rapidly added force, providing substantial support to prevent excessive trunk flexion, leading to improvements in WBAM. The minimum MOS evaluation showed that the Passive Mode caused detrimental effects, while the Active Mode showed potential for improving minimum MOS. Overall, the insignificant improvements or detrimental effects of the Passive Mode are aligned with previous studies. 

Though the study showed improvements in stability with the Active Mode, it presents a few limitations. Firstly, the sample size was limited to five subjects, which may limit generalizability. Further, we only evaluated a single perturbation magnitude. Testing the effectiveness of the device over multiple perturbation magnitudes in the future would further establish the efficacy of the device. Also, the IPAM was fully deflated (applying the maximum possible force) following the onset of perturbation for all participants. This force can be tuned by partially deflating the IPAM. Such tuning for individual subjects can further enhance the device's efficacy.  Lastly, unlike treadmill perturbations, overground stepping involves forward progression, and participants might have used tactics that combined treadmill motion with natural balance recovery \cite{tielke_non-ideal_2019}. 


\section{Conclusion}
This paper evaluated the effect of a lightweight semi-active BSD on stability following a standing perturbation. An active element in parallel with a passive BSD can rapidly provide assistive force upon the onset of a perturbation. This rapidly increasing force facilitated improvements in stability, quantified by the WBAM RMS and range, and the minimum MOS. In contrast, the device was ineffective without the parallel active element (Passive Mode), further supporting the semi-active device design. Though the sample size was small, the results demonstrate the semi-active BSD's potential for improving stability, and future validation with a larger sample size would further establish this potential.


\bibliographystyle{IEEEtran}
\bibliography{IROS26}
%


\end{document}